\documentclass[letterpaper, 10 pt, conference]{ieeeconf}  

\IEEEoverridecommandlockouts                              

\usepackage{dsfont}
\usepackage{amsmath,amssymb}
\usepackage{algorithm}
\usepackage{algpseudocode}

\DeclareMathOperator*{\argmax}{argmax} 
\let\labelindent\relax
\usepackage{enumitem}
\usepackage{subcaption}
\usepackage{graphicx}
\usepackage{booktabs}
\usepackage{multirow}
\usepackage{caption} 
\usepackage{hyperref}

\title{\LARGE \bf
Robot Policy Transfer with Online Demonstrations: An Active Reinforcement Learning Approach
}

\author{Muhan Hou$^{1}$, Koen Hindriks$^{1}$, A.E. Eiben$^{1}$, Kim Baraka$^{1}$
\thanks{$^{1}$All authors are with the Department of Computer Science, Vrije Universiteit (VU) Amsterdam, The Netherlands, and can be contacted at {\tt\small m.hou@vu.nl}}%
}

\begin{document}

\maketitle
\thispagestyle{empty}
\pagestyle{empty}

\begin{abstract}
Transfer Learning (TL) is a powerful tool that enables robots to transfer learned policies across different environments, tasks, or embodiments. To further facilitate this process, efforts have been made to combine it with Learning from Demonstrations (LfD) for more flexible and efficient policy transfer. However, these approaches are almost exclusively limited to \textit{offline demonstrations} collected before policy transfer starts, which may suffer from the intrinsic issue of covariance shift brought by LfD and harm the performance of policy transfer. Meanwhile, extensive work in the learning-from-scratch setting has shown that \textit{online demonstrations} can effectively alleviate covariance shift and lead to better policy performance with improved sample efficiency. This work combines these insights to introduce \textit{online demonstrations} into a policy transfer setting. We present Policy Transfer with Online Demonstrations, an active LfD algorithm for policy transfer that can optimize the timing and content of queries for online episodic expert demonstrations under a limited demonstration budget. We evaluate our method in eight robotic scenarios, involving policy transfer across diverse environment characteristics, task objectives, and robotic embodiments, with the aim to transfer a trained policy from a source task to a related but different target task. The results show that our method significantly outperforms all baselines in terms of average success rate and sample efficiency, compared to two canonical LfD methods with offline demonstrations and one active LfD method with online demonstrations. Additionally, we conduct preliminary sim-to-real tests of the transferred policy on three transfer scenarios in the real-world environment, demonstrating the policy effectiveness on a real robot manipulator.
\end{abstract}


\section{Introduction}

Transfer Learning (TL) has recently become increasingly popular in the robotics community. Instead of learning from scratch, TL enables a robot to utilize prior knowledge (e.g., policy) of a source situation to improve learning in a related target situation \cite{jaquier2023transfer}. This target situation may differ from the source in various aspects, such as environments \cite{zhang2023efficient, ren2023adaptsim} (e.g., from simulation to real world), tasks \cite{hundt2021good} (e.g., grabbing a bottle to grabbing a mug), or even embodiments \cite{jian2023policy} (e.g., different robot arms).

To facilitate policy transfer, efforts have been made to combine TL with Learning from Demonstrations (LfD).
These demonstrations can be utilized to acquire prior skills of the source domain \cite{wang2018transferring, fitzgerald2014representing, akbulut2021acnmp, pertsch2022cross, gao2023transferring} and accelerate the policy transfer process to the target domain \cite{yao2021integrating, hundt2021good}. However, almost all previous work relies on \textit{offline demonstrations}, which are collected before policy transfer starts.  This can lead to the covariance shift \cite{ross2010efficient}, as the state distribution in the demonstrations differs from that of the ongoing robot policy. This issue can even be further exacerbated by discrepancies in task space between the source and target domains.
And to adapt the demonstration distribution to the ongoing policy update, it requires information about the learning process, which is not available a prior and can only be obtained iteratively in an online manner. By contrast, extensive work \cite{zhang2016query, chen2020active, rigter2020framework} in the learning-from-scratch setting has shown that \textit{online demonstrations}, which are collected during the learning process, can effectively alleviate covariance shift and produce better policy performance. Furthermore, by querying for the most suitable demonstrations during the learning process, LfD with online demonstrations can achieve the same level of policy performance with much fewer demonstrations \cite{sakr2023can} and improve sample efficiency \cite{rigter2020framework}. This can be of great importance especially when the number of demonstrations is constrained by time and financial costs (e.g., data collection from human demonstrators). Therefore, we extend this idea to the scenario of policy transfer and ask the question: \textit{will online demonstrations benefit policy transfer, and how can we realize that?}

\begin{figure*}[t!]
  \centering
  \includegraphics[width=0.9\linewidth]{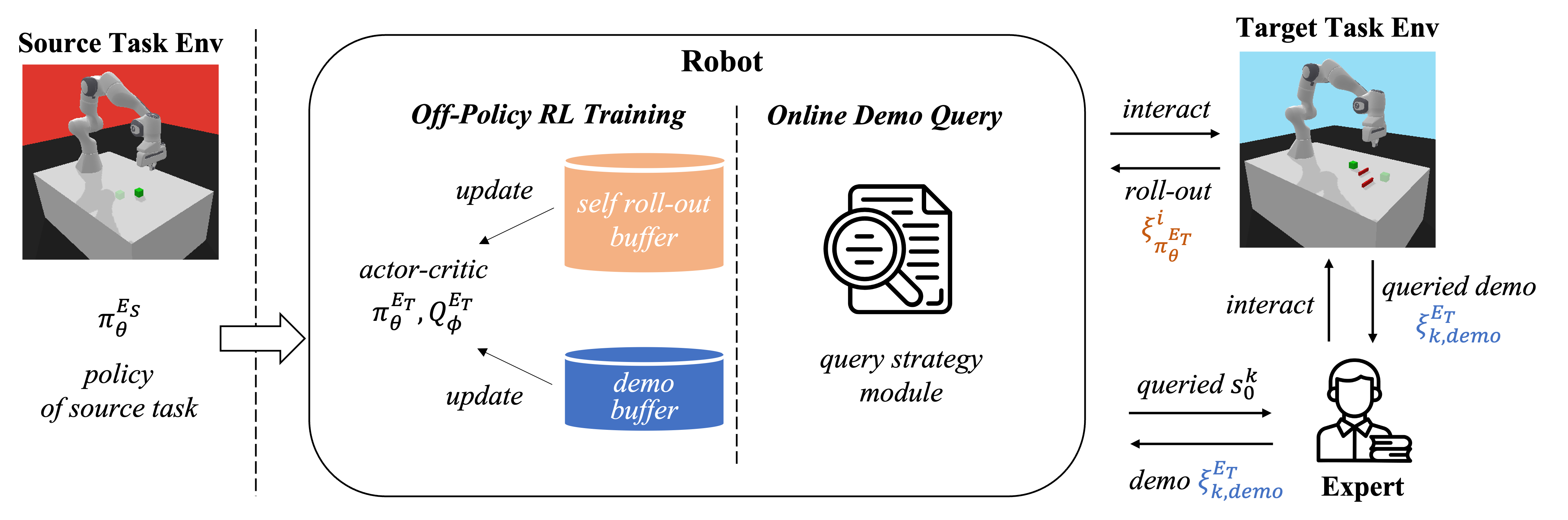}
  \caption{Overview of our method.}
  \label{overview}
  \vspace{-5mm}
\end{figure*}

To this end, we present the first active LfD algorithm for policy transfer that leverages online demonstrations for transfer learning and show that online demonstrations \textit{will} indeed benefit policy transfer if designed properly. Our method extends the EARLY framework \cite{hou2024give} to the transfer learning scenario and simultaneously optimizes the problems of \textit{when} to query and \textit{what} episodic demonstrations of the target task to query during the policy transfer process (see figure \ref{overview}). More specifically, a query strategy module evaluates the trajectory-based uncertainty after each episodic roll-out of the robot policy for the target task. An adaptive threshold of uncertainty is then updated to determine whether a query will be requested. If so, the query strategy module outputs the initial state of one of the recent roll-outs that the robot is most uncertain about and queries an episodic expert demonstration starting from this initial state. We test our approach in $8$ robot policy transfer scenarios, involving policy transfer across diverse environment characteristics, task objectives, and robotic embodiments . Compared to EARLY and two other canonical LfD baselines with offline demonstrations, our method achieves significantly higher success rate and sample efficiency in target tasks that the other baseline methods can hardly address. To summarize, our contributions are two-fold:
\begin{enumerate}[leftmargin=*]
    \item We present an active LfD algorithm for robot policy transfer that effectively transfers policies from source to target tasks across diverse scenarios. 
    \item We validate the effectiveness of our method through a set of robotic policy transfer scenarios, supported by extensive empirical results from simulations and preliminary sim-to-real experiments in real-world environments.
\end{enumerate}


\section{Related Work}

\textbf{Policy Transfer with Demonstrations}
Policy transfer aims to adapt a policy of a source domain to a target domain that is related but different from the source \cite{jaquier2023transfer}. To further facilitate policy transfer, efforts have been made to introduce demonstrations into this process, either from the source domain or the target domain. When these demonstrations are from the source domain, they are commonly used to acquire prior skills of the source domain using different representations \cite{wang2018transferring, fitzgerald2014representing, akbulut2021acnmp, pertsch2022cross, gao2023transferring}, whcih can be later fine-tuned for the taregt domain. Additionally, these source domain demonstrations can also be used to infer the dynamics difference between the target and source domain, the knowledge of which can be then utilized to adapt policy to the target domains \cite{kim2020domain}. And when these demonstrations are from target domain, they can be utilized to accelerate the policy transfer process \cite{yao2021integrating, hundt2021good} and better transfer the robot policy from simulation to real-world environments \cite{desai2020imitation}. However, almost all of these work only use offline demonstrations, which are collected prior to the policy transfer process. By contrast, extensive evidence has shown in the scenario of learning from scratch that online demonstrations can effectively improve sample efficiency and achieve better learning performance by providing the required demonstrations to the learning agent during the training process. Therefore, our work extends this idea to the scenario of policy transfer and presents an active LfD algorithm that queries online demonstrations, achieving better performance in policy transfer with improved sample effiency.


\textbf{Interactive Learning from Demonstrations}. To alleviate the issue of covariance shift of conventional Learning from Demonstrations (LfD), efforts have been made to introduce interactive teaching inputs into the learning paradigm \cite{hou2023shaping, 10309326}. Instead of using offline demonstrations that are collected before learning happens, interactive LfD collect teaching signals from demonstrators during the learning process. These teaching inputs can be in the form of evaluative feedback \cite{najar2020interactively, griffith2013policy,  knox2008tamer, biyik2022learning, cui2021empathic}, state-action corrections \cite{perez2020interactive, chisari2022correct, celemin2023knowledge, celemin2019interactive}, and episodic demonstrations \cite{rigter2020framework, hou2024give}, the timing and content of which can either be determined by the demonstrators themselves \cite{ross2010efficient, kelly2019hg} or actively requested by the learning agent \cite{zhang2016query, chen2020active, rigter2020framework, hou2024give}. However, all of them are designed for the scenario of learning from scratch. Few have attempted to explore its potential in policy transfer, which is the main focus of our work.


\section{Background}
To leverage \textit{online} demonstrations for robot policy transfer, our work builds upon the Advantage Weighted Actor-Critic (AWAC), an advanced off-policy LfD to update actor-critic models with \textit{offline} demonstrations. Furthermore, we optimize both the timing and content of online demonstration queries for \textit{policy transfer} by adapting the general paradigm for active LfD proposed in the EARLY framework \cite{hou2024give} that was originally designed for the scenario of \textit{learning from scratch}. We thereby introduce the necessary background as follows.

\subsection{Advantage Weighted Actor-Critic} \label{AWAC}
Advantage Weighted Actor-Critic (AWAC) \cite{nair2020awac} is an advanced LfD algorithm that optimizes the policy and value function networks with offline demonstrations. It utilizes the Soft Actor-Critic (SAC) \cite{haarnoja2018soft} as its underlying off-policy RL algorithm and updates its action-value function $Q_{\phi}(s, a)$ via:
\begin{equation}\label{q}
    \phi_k = \arg\min_{\phi} \mathbb{E}_{D} \left[ (Q_{\phi}(s, a) - y)^2 \right],
\end{equation}
where
\begin{equation}
    y = r(s,a) + \gamma \mathbb{E}_{s', a'} \left[ Q_{\phi_{k-1}(s', a')} \right].
\end{equation}
And for the policy network $\pi_{\theta}$, AWAC updates its parameters $\theta$ via:
\begin{equation}\label{pi}
    \theta_{k+1} = \arg\max_{\theta} \mathbb{E}_{s,a \sim \beta} \left[\log\pi_{\theta}(s|a) \exp \left( \frac{1}{\lambda} A^{\pi_{\theta}}(s,a) \right) \right],
\end{equation}
where $\beta = \mathcal{D} \cup \mathcal{R}$ is the combined replay buffer consisting of the demo buffer $\mathcal{D}$ and self roll-out buffer $\mathcal{R}$. $\lambda$ is a hyperparameter and $A^{\pi_{\theta}}(s,a) = Q^{\pi_{\theta}}(s,a) - V^{\pi_{\theta}}(s)$ is the advantage for the policy $\pi_{\theta}$, which is used to bias the policy update to favor actions of larger estimated action-values.


\subsection{EARLY} \label{EARLY}
Episodic Active Reinforcement Learning from demonstration querY (EARLY) \cite{hou2024give} is an active LfD algorithm that leverages online demonstrations for the scenario of learning from scratch. It decides \textit{when} to query and \textit{what} episodic expert demonstration to query by evaluating the uncertainty $u$ of an episodic roll-out $\xi_{\pi}^i$ under policy $\pi$ with:
\begin{equation} \label{uncertainty eq}
     u(\xi_{\pi}^i) = \mathbb{E}_{(s_t^i, a_t^i) \in \xi_{\pi}^i} \left[ | r_{t}^i + Q(s_{t+1}^i, a_{t+1}^i) - Q(s_t^i, a_t^i)| \right].
\end{equation}
After each episodic roll-out, EARLY will update a shifting history of uncertainty to generate an adaptive threshold to decide whether to query. Once deciding to query, it will request an episodic expert demonstration starting from a certain initial state chosen by the query strategy. Both policy roll-outs and expert demonstrations will be stored in the replay buffer, which will be used to update the actor and critic with SAC as the underlying off-policy RL algorithm.


\section{Methodology}

\subsection{Problem Formulation}
We consider the scenario where there is a source task environment $E_S$ and a target task environment $E_T$. Both environments are Markov Decision Processes (MDPs), denoted as $E_S = \{S^{E_S}, A^{E_S}, r^{E_S}, P^{E_S}, \rho^{E_S}_0, \gamma^{E_S}\}$ and $E_T = \{S^{E_T}, A^{E_T}, r^{E_T}, P^{E_T}, \rho^{E_T}_0, \gamma^{E_T}\}$, where $S$ is the state space, $A$ is the action space, $r: S \times A \rightarrow \mathbb{R}$ is the reward function, $P(s_{t+1} | s_t, a_t)$ is the environment transition function, $\rho_0$ is the initial state distribution, and $\gamma$ is the discount factor. 

For the source task $E_S$, we assume that the robot has access to the trained expert-level policy network $\pi_{\theta}^{E_S}$ parametrized by $\theta$. For the target task $E_T$, the robot has access to $N_d$ episodic expert demonstrations $\xi^{E_T}_{k, demo}=\{(s^{E_T}_{0,k}, a^{E_T}_{0,k}, r^{E_T}_{0,k}, s^{E_T}_{1,k}), ...\}$. Note that each demonstration can be queried from an expert demonstrator and iteratively added to the demo buffer $\mathcal{D}$ either before policy transfer starts (i.e., offline demonstrations) or during the training process (i.e., online demonstrations). The goal is to leverage these demonstrations to adapt the policy $\pi_{\theta}^{E_S}$ of the source task $E_S$ to the target task $E_T$ such that the transferred policy $\pi_{\theta}^{E_T}$ may maximize the discounted target task reward $J^{E_T} = \mathbb{E}_{\pi_{\theta}^{E_T}, s^{E_T}_0 \sim \rho^{E_T}_0}\left[ \sum_{t=0}^{L^{E_T} - 1} \left(\gamma^{E_T}\right)^t r^{E_T}(s^{E_T}_t, \pi_{\theta}^{E_T}(s^{E_T}_t)) \right]$ for an episode of the maximum length $L^{E_T}$. 

To simplify the problem, this work focuses on the scenario where the target and source task environments share the same state and action spaces but may differ in their transition dynamics $P$ and reward functions $r$. Additionally, we assume that the policy networks for both the source and target tasks share the same neural network structure.

\subsection{Policy Transfer With Online Demonstrations}
Aiming to better transfer the robot policy of a source task to a target task, we design an algorithm that learns in a ``demo-while-training'' manner, where the robot actively decides the timing and contents of online expert demonstrations to query for better policy transfer performance. 

Starting from a trained policy $\pi_{\theta}^{E_S}$ of the source task, our algorithm evaluates the uncertainty $u$ using \eqref{uncertainty eq} after each episodic roll-out in the target task and updates a shifting history $H$ of roll-out initial states and a shifting history $H_u$ of roll-out uncertainty. After the shifting history grows to its full length $N_h$, the algorithm will generate an adaptive threshold $u_{thres}$ of uncertainty to determine whether to query. If it decides to query, the algorithm will search the shifting history $H$ for the episodic roll-out that has the highest uncertainty, and query an expert demonstrator for an episodic demonstration $\xi^{E_T}_{k, demo}$ in the target task starting from the same initial state as that of the most uncertain roll-out. We summarize the pseudo-code as in Algorithm \ref{alg}. 

\begin{algorithm}[t!]
\caption{Policy Transfer with Online Demonstrations}\label{alg}
\begin{algorithmic}[1]
\Require demonstration budget $N_d$, max length of shifting history $N_h$, ratio threshold $r_{query}$

\State Randomly initialize action-value network $Q_{\phi}^{E_T}$ for the target task
\State Initialize policy network $\pi_{\theta}^{E_T} \gets \pi_{\theta}^{E_S}$
\State Initialize self roll-out buffer $\mathcal{R}$ and demo buffer $\mathcal{D}$
\State Initialize initial state history $H$, roll-out uncertainty history $H_u$
\State Index of the adaptive threshold $idx_{thres} \gets int(N_h \times r_{query})$
\State Queried demo number $n_d \gets 0$
\For{iteration $i \in \{1, 2, ...\}$}
    \State roll out the policy $\pi_{\theta}^{E_T}$ to get a trajectory $\xi^i_{\pi_{\theta}^{E_T}}$
    \For{step $t \in \xi^i_{\pi^{E_T}}$}
        \State update self roll-out buffer $\mathcal{R}$
        \State update $Q_{\phi}^{E_T}$ and $\pi_{\theta}^{E_T}$ via \eqref{q} and \eqref{pi} respectively
    \EndFor
    \State calculate roll-out uncertainty $u_i(\xi^i_{\pi^{E_T}})$ via \eqref{uncertainty eq}
    \State update $H$ and $H_u$
    \If{l$en(H)$ $\geq$  $N_h + 1$}
        \State descendingly reorder $H_u$ into $H_u^{dsc}$
        \State adaptive threshold $u_{thres} \gets H_u^{dsc}[idx_{thres}]$
         \If{$u_i > u_{thres}$ and $n_d < N_d$}
             \State $s^{E_T}_{0,query} \gets \argmax_{s^{E_T}_{0, j} \in H}H_u$
             \State query a demo $\xi^{E_T}_{k, demo}$ starting from $s^{E_T}_{0,query}$
             \State update demo buffer $\mathcal{D}$
             \State update $Q_{\phi}^{E_T}$ and $\pi_{\theta}^{E_T}$
             \State $n_d \gets n_d + 1$
        \EndIf
        \State remove the earliest element from $H$ and $H_u$
    \EndIf
\EndFor
\end{algorithmic}
\label{algorithm1}
\end{algorithm}
	
\section{Experimental Setup}

\begin{figure*}[t!]
  \centering
  \includegraphics[width=0.9\linewidth]{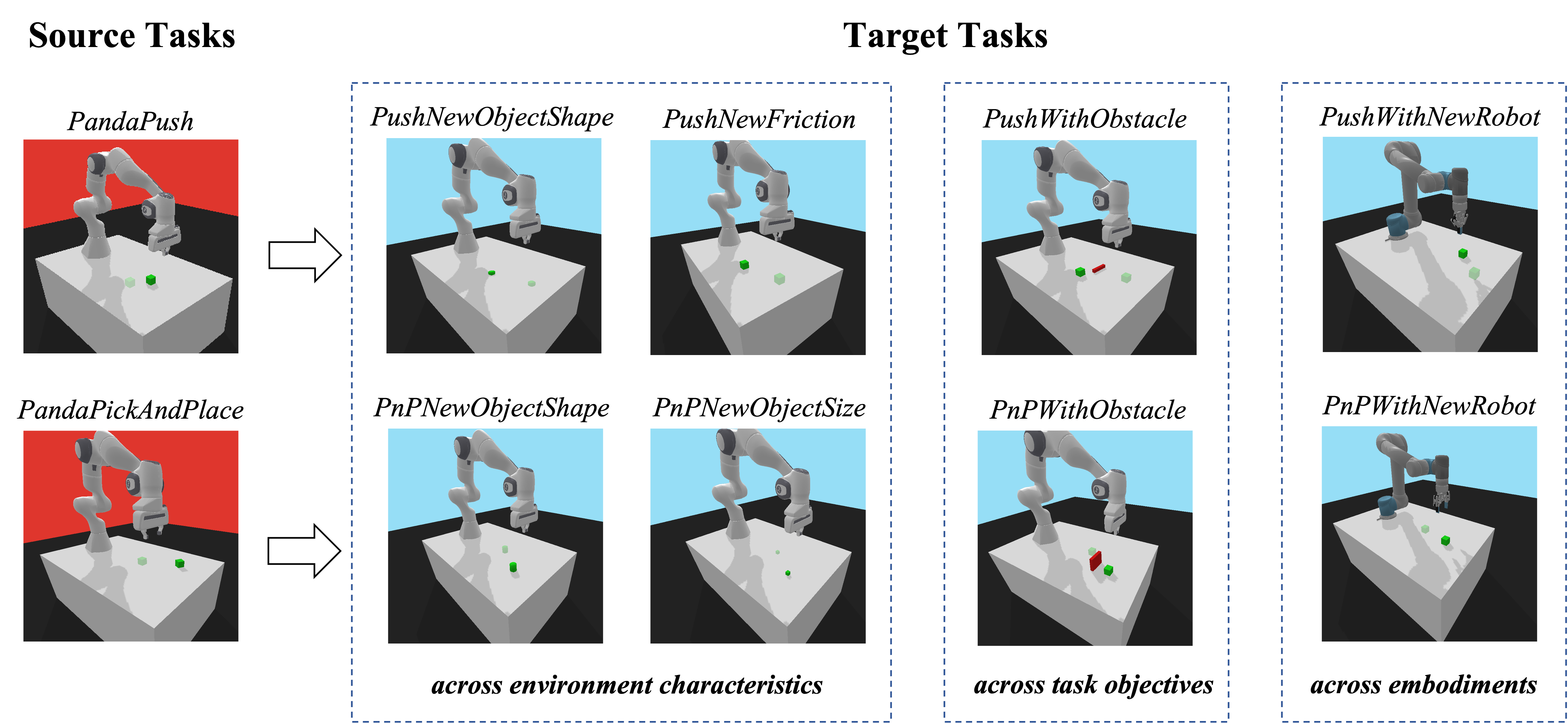}
  \caption{Overview of all the source and target tasks.}
  \label{target environments}
  \vspace{-5mm}
\end{figure*}

\subsection{Task Environments}
For source tasks, we choose two robotic tasks: \textit{PandaPush} and \textit{PandaPickAndPlace} as designed in~\cite{gallouedec2021pandagym}. For the task of \textit{PandaPush}, the Franka Emika manipulator, which is of $7$ degrees of freedom, starts from a neutral pose and aims to push a $0.04$m-sized cube across the table surface to a goal position randomly chosen from an area of $0.3$m $\times$ $0.3$m. For the task of \textit{PandaPickAndPlace}, the Franka Emika manipulator starts from the same neutral pose and aims to pick up a cube positioned in an area of $0.3$m $\times$ $0.04$m and place it at a pre-defined fixed position on the table surface. Both source tasks are situated in an environment without any obstacles on the table. Furthermore, both tasks are designed with sparse rewards, i.e., the robot receives a reward of $+1000$ only if the cube reaches the goal position, and a reward of $-1$ otherwise.

\begin{figure*}
  \centering
  \includegraphics[width=1.0\linewidth]{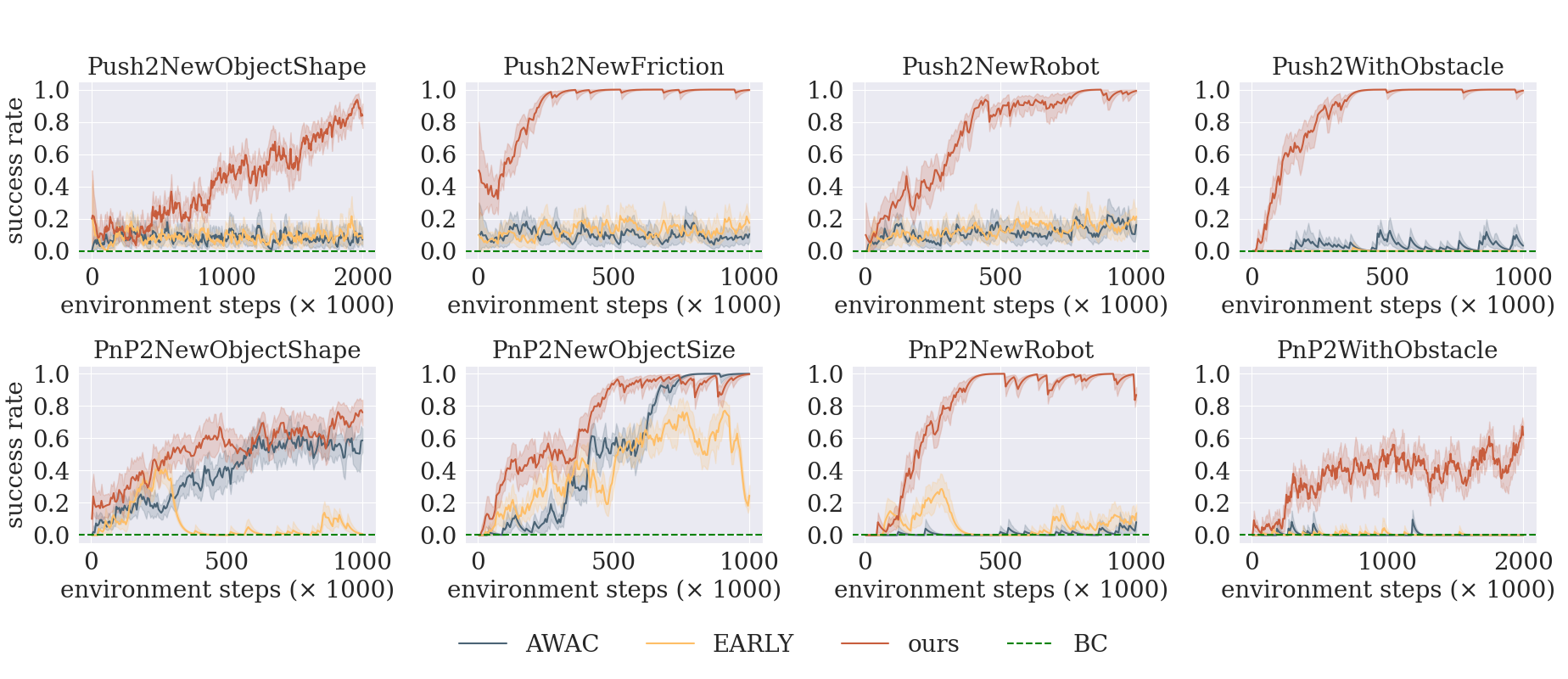}
  \caption{Results of simulation experiments.}
  \label{simulation results}
\end{figure*}

For target tasks, we design $8$ target tasks, as shown in Figure \ref{target environments}. More specifically, these target tasks can be divided into $3$ general types of transfer learning scenarios, i.e., across different \textit{environment characteristics} (e.g., object size, object shape, friction factor), across different \textit{task objectives} (e.g., pushing or pick-and-place a cube while avoiding obstacles), and across different \textit{robotic embodiments} (e.g., transfer the policy from one robot arm to a different one). 

Across different environment characteristics, we design $4$ target tasks:
\begin{itemize}
    \item \textit{PushNewObjectShape}: This task is similar to \textit{PandaPush}, but uses a coin-shaped object, which is a cylinder with a height of $0.01$m and a radius of $0.02$m.
    \item \textit{PushNewFriction}: This task is similar to \textit{PandaPush}, but features a friction factor of $0.25$, which is half the value used in \textit{PandaPush}.
    \item \textit{PnPNewObjectShape}: This task is similar to \textit{PandaPickAndPlace}, but uses a cylinder-shaped object with a height of $0.04$m and a radius of $0.02$m
    \item \textit{PnPNewObjectSize}: This task is similar to \textit{PandaPickAndPlace}, but uses a much smaller cube with a size of $0.02$m.
\end{itemize}

Across different task objectives, we designed $2$ more target tasks:
\begin{itemize}
    \item \textit{PushWithObstacle}: This task is similar to \textit{PandaPush}, but the robot arm needs to push the cube to a fixed goal position without touching the obstacle. An episode will terminate with a reward of $-1000$ once the cube touches the obstacle. The initial position of the cube will be randomly chosen from a fixed area of $0.3$m $\times$ $0.04$m.
    \item \textit{PnPWithObstacle}. This task is similar to \textit{PandaPickAndPlace}, but the robot arm needs to pick and place the cube to a fixed goal position without touching the obstacle. An episode will terminate with a reward of $-1000$ once the cube runs into the obstacle. The initial position of the cube will be randomly chosen from a fixed area of $0.3$m $\times$ $0.04$m.
\end{itemize}

And across different robotic embodiments, we design another $2$ target tasks:
\begin{itemize}
    \item \textit{PushNewRobot}: This task is similar to \textit{PandaPush}, but uses UR5 as the robot manipulator, which is of $6$ degrees of freedom. Since we use the target 3D position of the end-effector as the action $a_t$, switching to a new robot of a different number of degrees of freedom will not lead to the mismatch of action space between the source task and the target task.
    \item \textit{PnPNewRobot}: This task is similar to \textit{PandaPickAndPlace}, but also uses UR5 as the manipulator instead of the Franka Emika manipulator.
\end{itemize}

Based on these $8$ target tasks, we define $8$ corresponding scenarios for robot policy transfer. For PushNewObjectShape, PushNewFriction, PushWithObstacle, and PushNewRobot, we use PandaPush as the source task, resulting in $4$ scenarios that we refer to as \textit{Push2NewObjectShape}, \textit{Push2NewFriction}, \textit{Push2WithObstacle}, and \textit{Push2NewRobot}. Similarly, we choose PandaPickAndPlace as the source task for PnPNewObjectShape, PnPNewObjectSize, PnPWithObstacle, and PnPNewRobot, resulting in another $4$ scenarios we refer to as \textit{PnP2NewObjectShape}, \textit{PnP2NewObjectSize}, \textit{PnP2WithObstacle}, and \textit{PnP2NewRobot}.

\subsection{Demonstration Collection}
For each of the $8$ scenarios for policy transfer, we use a joystick to collect a pool of $30$ episodic demonstrations from the same human expert demonstrator and guarantee that each demonstration succeeds in the target task. The initial states of these demonstrations are decided by uniformly sampling from the initial state distribution of the corresponding target task. During the learning process, the robot may iteratively request $10$ demonstrations (i.e., demonstration budget $N_d = 20$) from the pool. When an online query (i.e., $s^{E_T}_{0,query}$) is generated, we will choose the demonstration $\xi^{E_T}_{k, demo}$ from the pool whose initial state is closest to the queried one based on their Euclidean distance in the state space, and use this demonstration as the queried online demonstration. 

It should be noted that the distribution of demonstrations in the pool is unknown to the robot throughout the training process. It is only a methodological trick to function as an unknown human-expert policy. It iteratively outputs human demonstrations and feeds them back to the robot upon request. How each demonstration is retrieved from the pool is also unknown to the robot. This is equivalent to having a real human expert available throughout the training phase, iteratively providing demonstrations with the joystick whenever the robot queries.

\subsection{Baselines}
We compare our method with $3$ other popular LfD baselines in the scenario of policy transfer, including $2$ canonical LfD methods using offline demonstrations and $1$ active LfD method using online demonstrations. Each baseline is initialized with the same trained expert-level policy network of the source task and may request $10$ episodic expert demonstrations for the corresponding target task, either iteratively during the training process (i.e., online demonstrations) or all before the training process (i.e., offline demonstrations).

More specifically, we compare our method with $3$ other baselines as follows:
\begin{itemize}
    \item \textbf{AWAC}. An advanced LfD method using offline demonstrations as mentioned in Section \ref{AWAC}. For each transfer scenario, we uniformly sample $10$ episodic demonstrations from the corresponding demonstration pool and load them into the demo buffer before the training starts. We employ the same hyperparameter values for policy training as those specified in the original work.
    \item \textbf{Behavioral Cloning (BC)}. A common offline LfD algorithm that trains the policy via supervised learning \cite{pomerleau1988alvinn}. We sample the offline demonstrations in the same way as for the baseline of AWAC.
    \item \textbf{EARLY}. An active LfD method using online demonstrations as mentioned in Section \ref{EARLY}. When a query is generated by the method, we choose the expert demonstration from the demonstration pool whose initial state is closest to the request one and feed it to the algorithm. All the hyperparemeters are set as the same values as in the original work.
\end{itemize}

For our method, we set the max length of shifting history $N_h = 20$ and the ratio threshold $r_{query}=0.1$ for all $8$ transfer scenarios. We obtain the trained expert-level policies for the source tasks by training the actor-critic models via SAC using the dense reward function shaped by the distance to the goal. We employ the same actor-critic model structure as AWAC with the default hyperparameters as in \cite{nair2020awac}. Furthermore, we add layer normalization for more stable convergence as in \cite{ball2023efficient}. We also use a balanced-sampling strategy as EARLY for each policy and action-value update by sampling an equal number of transitions from both the demo buffer $\mathcal{D}$ and self roll-out buffer $\mathcal{R}$.


\section{Results and Discussion}

\subsection{Simulation Experiments}

\textbf{Higher success rate with improved sample efficiency.} \
We evaluate each baseline method by measuring the success rate of the target task policy every 5000 environment steps, using 10 random seeds. As shown in Figure \ref{simulation results}, our method consistently outperforms all baselines in terms of average success rate across all $8$ policy transfer scenarios. Specifically, our method achieves around $100\%$ average success rate in $6$ scenarios: \textit{Push2NewObjectShape}, \textit{Push2NewFriction}, \textit{Push2NewRobot}, \textit{Push2WithObstacle}, \textit{PnP2NewObjectSize}, and \textit{PnP2NewRobot}. In contrast, the other baselines struggle to transfer policies effectively, with average success rates generally not exceeding $20\%$, except in the \textit{PnP2NewObjectSize} scenario. In that case, while EARLY reaches an $80\%$ success rate, its performance quickly degrades and fails to converge. Although AWAC converges to a $100\%$ success rate, it requires approximately $750 \times 10^3$ steps, converging abot $50\%$ slower than our method.

For the remaining $2$ scenarios, our method still performs better, achieving over $70\%$ average success rate in \textit{PnP2WithObstacle}, where other baselines struggle to solve the target tasks. In the \textit{PnP2NewObjectShape} scenario, our method also reaches around $80\%$ success, while BC and EARLY fail to solve the task, and AWAC converges to only around $60\%$.

These strong empirical results demonstrate that our method maintains robust effectiveness across a variety of transfer learning scenarios, successfully transferring policies from source tasks to target tasks across diverse environmental conditions, task objectives, and robotic embodiments.

\begin{figure}[t!]
\centering
\includegraphics[width=0.45\textwidth]{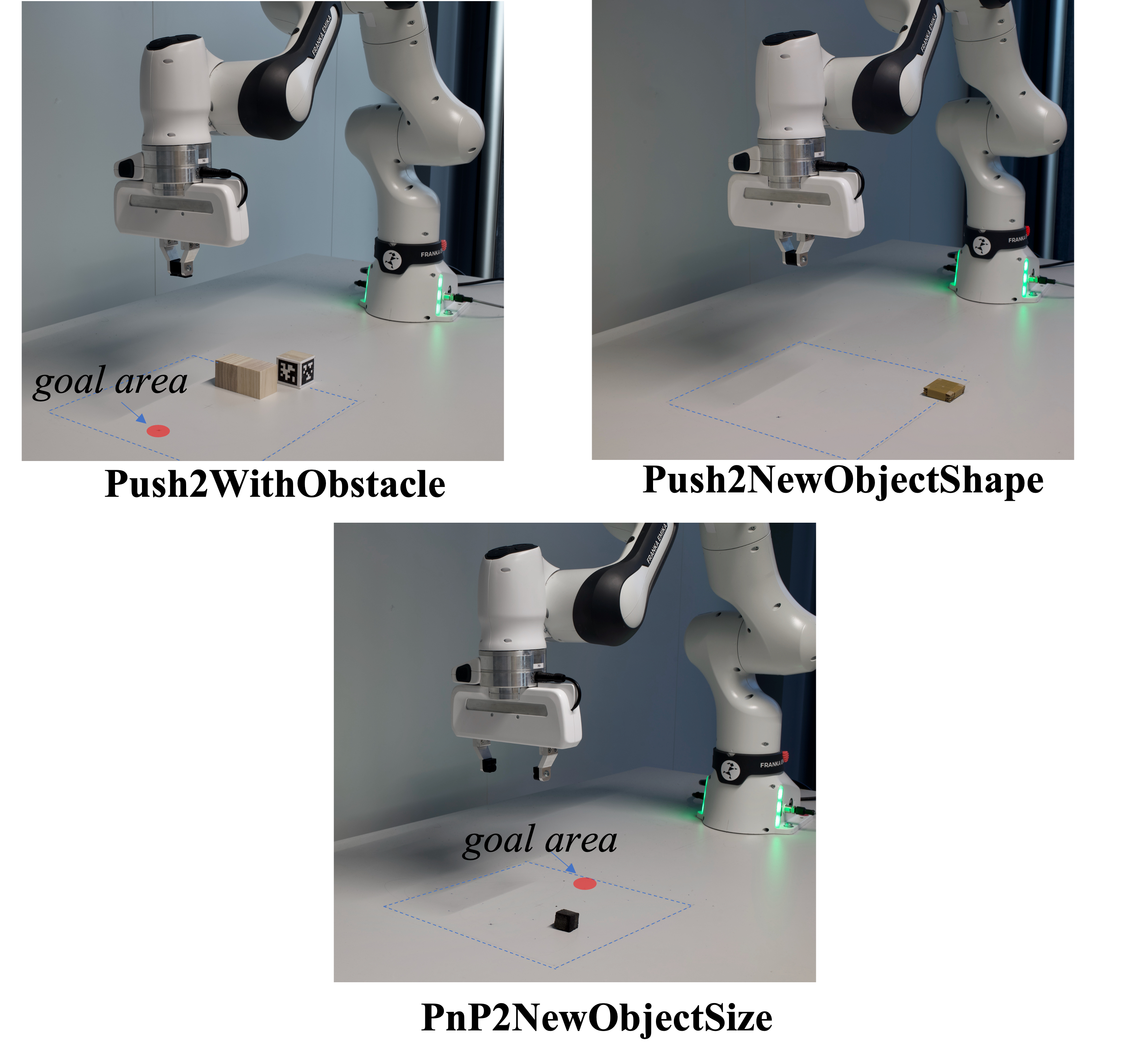}
\caption{The scenarios for real-world experiments.}
\label{real-world scenes}
\vspace{-5mm}
\end{figure}

\begin{table}[t!]
\caption{Results of real-world experiments.}
\label{real-world results}
  \begin{center}
    \begin{tabular}{ccccc}
      \toprule 
         \multirow{2}{*}{Scenario} & \multicolumn{4}{c}{Success Rate } \\ \cmidrule(lr){2-5}
                         & ours & AWAC & BC & EARLY \\
       \midrule
       Push2WithObstacle & \textbf{100\% } & 0\% & 0\% & 0\% \\
       Push2NewObjectShape & \textbf{80\% } & 0\% & 0\% & 0\% \\
       PnP2NewObjectSize & \textbf{50\% } & \textbf{50\% } & 0\% & $10\%$ \\
      \bottomrule 
    \end{tabular}
  \end{center}
  \vspace{-5mm}
\end{table}

\subsection{Preliminary Sim-to-Real Tests on the Real Robot}
We replicate the target task environments of PushWithObstacle, PushNewObjectShape, and PnPNewObjectSize in the real-world environment (shown in figure \ref{real-world scenes}) and validate the sim-to-real performance of our method on the real Franka Emika manipulator. More specifically, we use an $0.04$m-sized cube for PushWithObstacle and an obstacle with dimensions of ($0.02$m, $0.1$m, $0.02m$) located it at $(-0.11, 0, 0.01)$. For PushNewObjectShape, we use the a flat cylinder with $h=0.01$m and $r=0.02$m. For PnPNewObjectSize, we use a $0.02$m-sized cube. For both our method and the baselines, we evaluate their performance on the real robot by rolling out the transferred policies saved at the last checkpoint for $10$ test episodes. The initial state of each test episode is uniformly sampled from the initial state space. And to simplify the test, we assume that the object pose in the real world matches that in a synchronized simulation environment by using the real-time state in the simulation as the input to the trained policy. The trained policy outputs the real-time action defined as the target $3D$ position of the end-effector and sends it to the real robot under a frequency of $100$ Hz. Upon receiving the command, the robot converts it into the corresponding joint command and controls its joints with the joint-position controller implemented by the open-sourced library panda-py \cite{elsner2023taming}. As shown in table \ref{real-world results}, the transferred policies trained by our method achieve the highest success rate compared with other baselines, indicating the efficacy of the transferred policy obtained by our method in the real-world environment.


\section{Conclusions and Limitations}
In this work, we present an active LfD algorithm that introduces online demonstrations into the policy transfer process. By optimizing the timing and content of queries for online demonstrations, our method achieves significantly better performance for policy transfer with respect to both average success rate and sample efficiency in $8$ robotic scenarios compared with $3$ other LfD baselines. However, this work is only focused on scenarios where the target and source tasks share the same state-action space and policy network structure. This limitation may be alleviated if a more unified representation of task space and policy can be used as in \cite{jian2023policy}. Additionally, we assume access to an environment reset function that can reset the environment to a queried initial state, which may constrain its applicability in real-world settings, though this can be mitigated with human assistance or existing path-planning controllers. Furthermore, additional factors, such as user experience, need to be taken into account when online demonstrations are iteratively collected from real humans interacting with real robots. How to maintain good performance of policy transfer while accounting for human factors will be a key focus of our future work.

\clearpage


\bibliographystyle{IEEEtran}
\bibliography{example}

\end{document}